\documentclass[letterpaper, 10 pt, conference]{ieeeconf}  

\IEEEoverridecommandlockouts                              

\usepackage{url}
\usepackage{comment}
\usepackage{graphicx}
\usepackage{algorithm}
\usepackage{algorithmic}
\usepackage {threeparttable}
\usepackage{amsmath}
\usepackage{ascmac}
\usepackage{tcolorbox}
\usepackage{xcolor}
\usepackage{amssymb}
\usepackage{mathtools}
\usepackage{makecell}
\usepackage{xcolor} 
\usepackage{booktabs}
\usepackage{adjustbox}
\usepackage{cuted}
\usepackage{hyperref}

\title{\LARGE \bf
Memory Transfer Planning: LLM-driven Context-Aware Code Adaptation for Robot Manipulation
}

\author{Tomoyuki Kagaya*$^{1}$ Subramanian Lakshmi*$^{2}$ Yuxuan Lou$^{3}$ Thong Jing Yuan$^{2}$ Jayashree Karlekar$^{2}$ \\
Sugiri Pranata$^{2}$ Natsuki Murakami$^{1}$ Akira Kinose$^{1}$ Yang You$^{3}$
\thanks{*Equal contribution}
\thanks{$^{1}$Panasonic Connect Co., Ltd., Japan, $^{2}$Panasonic R\&D Center, Singapore, $^{3}$National University of Singapore, Singapore. Correspondence to: Tomoyuki Kagaya \texttt{<kagaya.tomoyuki@jp.panasonic.com>}}
}

\begin{document}

\maketitle
\thispagestyle{empty}
\pagestyle{empty}

\begin{abstract}

    Large language models (LLMs) are increasingly explored in robot manipulation, but many existing methods struggle to adapt to new environments. Many systems require either environment-specific policy training or depend on fixed prompts and single-shot code generation, leading to limited transferability and manual re-tuning. We introduce Memory Transfer Planning (MTP), a framework that leverages successful control-code examples from different environments as procedural knowledge, using them as in-context guidance for LLM-driven planning. Specifically, MTP (i) generates an initial plan and code using LLMs, (ii) retrieves relevant successful examples from a code memory, and (iii) contextually adapts the retrieved code to the target setting for re-planning without updating model parameters. We evaluate MTP on RLBench, CALVIN, and a physical robot, demonstrating effectiveness beyond simulation. Across these settings, MTP consistently improved success rate and adaptability compared with fixed-prompt code generation, naive retrieval, and memory-free re-planning. Furthermore, in hardware experiments, leveraging a memory constructed in simulation proved effective. MTP provides a practical approach that exploits procedural knowledge to realize robust LLM-based planning across diverse robotic manipulation scenarios, enhancing adaptability to novel environments and bridging simulation and real-world deployment.

\end{abstract}


\section{INTRODUCTION}

    Large Language Models (LLMs) are increasingly integrated into agent systems that combine planning and execution for robotic manipulation. Prior work has shown that LLMs can produce human-readable plans and executable control snippets, enabling new ways to specify and realize manipulation behaviors. However, adaptation to new environments remains difficult in practice. Many approaches rely on environment-specific policy training, limiting portability, while methods centered on fixed prompts or single-shot code generation often require manual prompt edits to maintain performance when the environment changes.

    We address this problem by viewing procedural knowledge from prior successful codes as a reusable resource for planning. We propose Memory Transfer Planning (MTP): when the initial plan is insufficient, MTP retrieves similar successful examples from a code memory and performs context-aware adaptation to the target setting, then supplies the adapted snippets as in-context guidance to the LLM for re-planning, all without updating model parameters.

    MTP comprises three components. (i) Code Generation: an LLM produces an initial plan and control code for the task. (ii) Memory Retrieval: a searchable memory stores successful examples together with task/context descriptors and retrieves relevant cases with task similarity. (iii) Re-planning with adapted memory: retrieved examples are adapted to the current context and integrated into the prompt for the next planning step.

    We validate MTP on RLBench \cite{james2019rlbenchrobotlearningbenchmark} and CALVIN \cite{mees2022calvinbenchmarklanguageconditionedpolicy} and on a physical manipulator. Across these settings, MTP consistently improves task success and adaptability relative to fixed-prompt generation, naïve retrieval, and memory-free re-planning. In ablations, both memory retrieval and context-aware adaptation prove necessary to realize the observed gains.

    Our contributions can be summarized as follows:

    \begin{itemize}
        \item We formulate MTP for robot manipulation: prior successful executable control code and traces are stored as procedural knowledge, then retrieved and context-aware adapted to guide an LLM planner, without parameter updates.

        \item We provide a plug-and-play loop (Code Generation, Memory Retrieval, and Re-planning with adapted memory), a code memory with similarity retrieval, context-aware code adaptation (retargeting, parameter scaling, pre/post-condition edits).

        \item We evaluate MTP on RLBench, CALVIN, and a real robot, showing consistent gains over fixed-prompt generation, naive retrieval, and memory-free re-planning. Relative to VoxPoser \cite{voxposer}, success rate improves from 39.3\% to 64.4\% on RLBench, from 52.0\% to 67.3\% on CALVIN, and from 30\% to 75\% on a real robot.

    \end{itemize}


\begin{figure*}[ht]
    \centering
    \includegraphics[width=\linewidth]{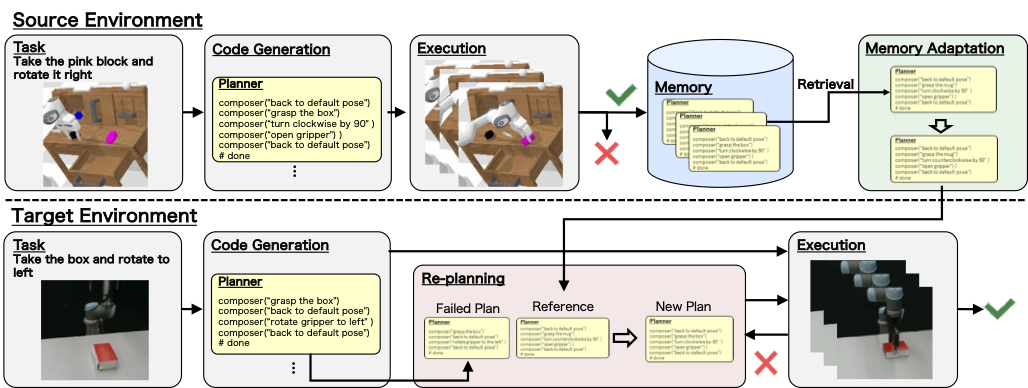}
    \caption{Overview of MTP. In the method for generating code for robot operation, the successfully generated code in the source environment is stored in memory and utilized for Re-planning in the target environment. At that time, codes with high query similarity are retrieved from memory and converted to match the style of the target environment, thereby facilitating smooth memory adaptation.}
    \label{fig:rlbench-pipeline}
\end{figure*}
\section{Related Work}
\label{sec:related_work}

\subsection{Advancements in Reasoning with LLMs}
    
    Recent progress in LLM reasoning focuses on improving problem-solving through advanced prompting and learning techniques. Chain-of-Thought prompting guides models to generate intermediate reasoning steps, enhancing performance on complex tasks \cite{wei2023chainofthoughtpromptingelicitsreasoning}. Tree-of-Thoughts methods extend this by exploring multiple reasoning paths and self-critiquing to find more robust solutions \cite{yao2023tree}.
    Incorporating external knowledge via Retrieval-Augmented Generation \cite{lewis2021retrievalaugmentedgenerationknowledgeintensivenlp} further improves factual accuracy and context. Reinforcement learning (RL) techniques have been applied to further enhance LLM reasoning. The RL-STaR \cite{chang2025rlstartheoreticalanalysisreinforcement} framework utilizes reinforcement learning to automatically generate reasoning steps, reducing reliance on human-labeled data and providing a theoretical foundation for understanding its effectiveness .

\subsection{LLM Agents with Memory}

    Memory modules enable LLM agents to accumulate and utilize knowledge through reading, writing, and reflection operations. Recent works like RAP \cite{kagaya2024rapretrievalaugmentedplanningcontextual} leverage past experiences for planning, while SayPlan \cite{rana2023sayplangroundinglargelanguage} uses scene graphs and environmental feedback as short-term memory for directing actions. HELPER \cite{sarch2023openendedinstructableembodiedagents} employs retrieval-augmented prompting over a memory of language–program pairs. The Generative Agent \cite{park2023generativeagentsinteractivesimulacra} implements both short-term contextual memory and long-term behavioral storage to guide decision-making. Research also explores the incorporation of a ``world model" into embodied agents, which allows them to understand and predict their environment through multimodal perception and memory \cite{fung2025embodiedaiagentsmodeling}.

\subsection{Embodied Agents}

    Recent advances in embodied AI focus on efficient planning and reasoning in physical environments. Hierarchical approaches \cite{sharma2022skillinductionplanninglatent} combine high-level LLM planning with low-level execution, while others focus on grounding actions within environments \cite{shinn2023reflexionlanguageagentsverbal}. Some methods address error recovery \cite{singh2022progpromptgeneratingsituatedrobot} or lifelong learning \cite{tziafas2024lifelongrobotlibrarylearning} showing promise in generating flexible robotic control without pre-training. However, these approaches still face challenges in generalizing across diverse environments. 
	

\begin{figure}[ht]
\centering
\begin{tcolorbox}[
  fontupper=\footnotesize\ttfamily\linespread{0.9}\selectfont,
  colback=gray!10, 
  colframe=black, 
  boxrule=0.5pt, 
  arc=2pt, 
  fonttitle=\bfseries,
  width=0.95\linewidth
]
[
  \{
    \textcolor{blue}{"environment"}: "RLBench",\\
    \textcolor{blue}{"query"}: "leave the pan open.",\\
    \textcolor{blue}{"code"}: \{\\
    \hspace*{2em}objects = ['saucepan', 'saucepan\_lid']\\
    \hspace*{2em}\# Query: leave the pan open.\\
    \hspace*{2em}composer("grasp the saucepan\_lid")\\
    \hspace*{2em}composer("move away from the saucepan by 25cm")\\
    \hspace*{2em}composer("open gripper")\\
    \hspace*{2em}composer("back to default pose")\\
    \hspace*{2em}\# done\\
    \},\\
    \textcolor{blue}{"status"}: "success"
  \},\\[0.5em]
  \{
    \textcolor{blue}{"environment"}: "RLBench",\\
    \textcolor{blue}{"query"}: "chuck way any rubbish on the table rubbish.",\\
    \textcolor{blue}{"code"}: \{\\
    \hspace*{2em}objects = ['bin', 'rubbish', 'tomato1', 'tomato2']\\
    \hspace*{2em}\# Query: chuck way any rubbish on the table rubbish.\\
    \hspace*{2em}composer("grasp the rubbish")\\
    \hspace*{2em}composer("back to default pose")\\
    \hspace*{2em}composer("move to the top of the bin")\\
    \hspace*{2em}composer("open gripper")\\
    \hspace*{2em}\# done\\
    \},\\
    \textcolor{blue}{"status"}: "success"
  \}
]
\end{tcolorbox}
\caption{Illustration of the format used to construct memory from successfully completed tasks, which allows the system to recall and utilize this memory during the execution of related tasks.}
\label{fig:memory_struct}
\end{figure}


\section{MTP: Memory Transfer Planning}
\label{sec:proposed_method}

    In this section, we introduce Memory Transfer Planning (MTP), a framework for reusing prior execution memories to guide Code Generation and Re-planning across tasks and environments.
    Fig. \ref{fig:rlbench-pipeline} provides an overview of the framework. The framework consists of three main components: Code Generation, Memory Retrieval, and Re-planning with transferred memory.
    
\subsection{Problem Formulation}

    In this work, we consider an agent operating in a particular environment $E$ and assigned with completing some task given a free-form language instruction $l$. To achieve this goal, the agent needs to generate robot trajectories $\tau_l$ based on the instruction $l$.

\begin{figure*}[h]
    \centering
    \includegraphics[width=\linewidth]
    {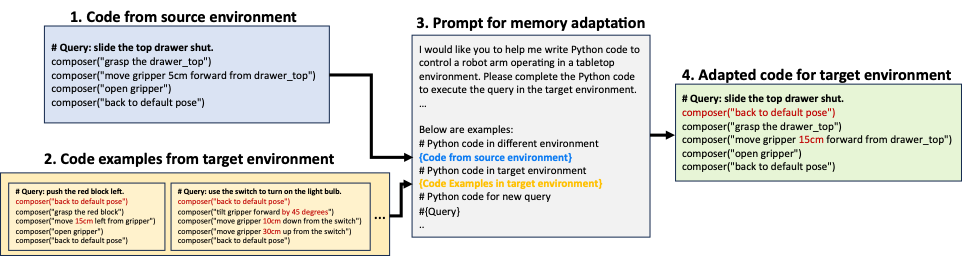}
    \vskip -0.1in
    \caption{The process flow of Memory adaptation.
    \textit{Prompt} uses \textit{code examples from the target environment (yellow block)} as a reference to convert the \textit{code from the source environment (blue block)} to generate the \textit{adapted code (green block)}.}
    \label{fig:kt_figure}
\end{figure*}

\subsection{Code Generation for Robot Manipulation}

    We use LLMs to generate Python code that is executed by an interpreter to decompose the language instruction into subtasks, invoke perception APIs and generate robot trajectories. 
    Our Code Generation pipeline follows the design from VoxPoser \cite{voxposer}, which comprises 3 LLM-calling steps:
    
    Step 1: The Planner is responsible for decomposing language-instruction guided tasks (e.g., ``press the light switch") into sub-tasks (e.g., ``grasp the button", ``move to the center of the button").  
    \begin{equation}
    \label{eqn:planner}
        Planner(l)=(l_1,l_2...,l_n)
    \end{equation}
    
    
    Step 2: The Composer takes in sub-task instruction $l_i$ (e.g., ``grasp the button", ``move to the center of the button") and invokes necessary low-level language model program (LMP). Each low-level LMP is responsible for unique functionality (e.g., parsing query objects in dictionary form, generating affordance map in Numpy array from Composer parametrization). 
    \begin{equation}
    \label{eqn:composer}
        Composer(l_i)=(LMP_1(l_i), LMP_2(l_i),.. LMP_k(l_i))
    \end{equation}
    
    Step 3: The low-level LMPs will be executed to interact with certain perception APIs to generate necessary value maps such as affordance and avoidance maps. These value maps will then be used together to find a sequence of end-effector positions serving as robot trajectories.
    \begin{equation}
    \label{eqn:lmp}
    exec_{LMP}(Composer(l_i))=\tau_{l_i}
    \end{equation}
    
    Thereafter, we combine the robot trajectories of each sub-task $l_i$ to form the complete robot trajectories attempting to accomplish the task.
    \begin{equation}
    \label{eqn:traj}
    \tau_l=(\tau_{l_1},\tau_{l_2}...,\tau_{l_n})
    \end{equation}
    

\begin{figure*}[h]
    \centering
    \includegraphics[width=\linewidth]{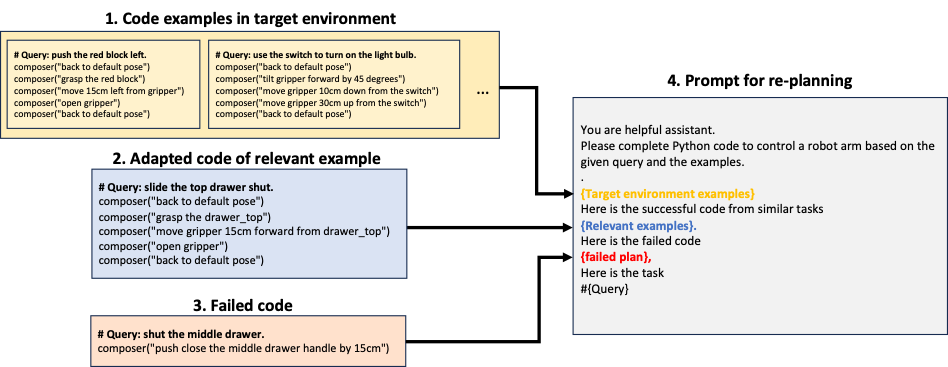}
    \vskip -0.1in
    \caption{Overview of \textit{prompt used for re-planning} based on \textit{target environment examples (yellow block)}, \textit{adapted code of the retrieved code from memory (blue block)}, and \textit{failed code (red block)}}
    \label{fig:replan_figure}
\end{figure*}


\subsection{Memory Construction and Retrieval}

Code generation is applicable across various environments, but often fails in unfamiliar scenarios.
In our method, task planning code generated by the Planner is stored in memory as procedural knowledge for future reuse. This enables the system to retrieve and adapt past successful examples based on context, allowing for flexible re-planning when the initial plan proves insufficient in a new environment.

Since the Composer and low-level LMPs typically exhibits more structured and task-invariant behavior, while the Planner must generate diverse and flexible outputs depending on task instructions, our memory module focuses exclusively on storing and retrieving Planner-level code.

\subsubsection{Memory Component}

    Here, we describe in detail the contents stored in the memory component shown in Fig. \ref{fig:rlbench-pipeline}.
    During the execution of tasks in the Code Generation pipeline, our system records the details of each task, including the environment, the input instructions, the generated code, and the outcomes (successes or failures).
    After execution and evaluation, only successfully executed codes are saved into our memory.

    Each successful log $L$ contains the following information: The environment E, the instruction $l$, and planner code $c$. Examples of the information recorded in memory are shown in Fig.  \ref{fig:memory_struct}, showcasing the types of data our system stores.

    \begin{equation}
    \label{eqn:planner_memory}  
    L=\{E, l, c\}
    \end{equation}

    \begin{algorithm}[tb]
       \caption{Re-planning in MTP}
       \label{alg:MTP}
    \begin{algorithmic}
       \STATE {\bfseries Initialize:} Environment $E$, Memory $M$, MaxTrial $d$
       \STATE {\bfseries Input:} Task $T$, Instruction $l$
       \STATE $c$ = CodeGeneration($l$, $E$)
       \STATE $result$ = Execution($c$, $E$)
       \IF{$result$ is Fail}
          \FOR{$i$ in range($d$ - 1)}
             \STATE $c_m$ = MemoryRetrieval($l$, $M$, $i$)    // Memory Retrieval for $i$th similar code
             \STATE $c_m'$ = MemoryAdaptation($m$)    // Memory Adaptation for target environment
             \STATE $c$ = CodeGeneration($l$, $c_m'$, $E$)    // Re-planning with adapted code
             \STATE $result$ = Execution($c$, $E$)
             \IF{$result$ is Success}
                \STATE break
             \ENDIF
          \ENDFOR
        \ENDIF
    \end{algorithmic}
    \end{algorithm}

\subsubsection{Memory Retrieval}

    When encountering a new task $T$, MTP can query the memory to find relevant successful logs.
    Assuming the instruction is $l$, for each log in the memory $L_{j}=\{E, l_j, c_j\}$, we will calculate a similarity score between the instructions from the memory log and the current instruction of task $T$.
    \begin{equation}
    \label{eqn:score}   
    Score(L_j) = cos\_sim(l, l_j)
    \end{equation}
    
    Here, the similarity score is calculated as the cosine similarity between the text embedding of current task instruction and logged instruction. Then we retrieve the codes with the top $k$ similarity scores. For generating these text embeddings, we employ sentence-transformers \cite{reimers2019sentencebert}.
    
    After retrieval, MTP transfers the retrieved codes and applies them as prompts for agents to generate more accurate code in the current task. This will be further explained in Section \ref{knwoledge_transfer_and_replanning}.
    
    \begin{equation}
    \label{eqn:retrieve}  
    \text{Retrieved}(c_{1:k}) = \text{TOP}_k^{\text{Score}(L_j)}\{c_j\}
    \end{equation}

\subsection{Memory Adaptation and Re-planning}
\label{knwoledge_transfer_and_replanning}

    In this section, we introduce two important designs to apply retrieved code from source environments to LLM code generation in the target environment: Memory adaptation and Re-planning.
    In Section \ref{knowledge_transfer}, we introduce how the Memory adaptation minimizes the difference in robotic manipulation code between source and target environments.
    In Section \ref{re_planning}, we introduce the instances and explain how the retrieved code is adapted to effectively leverage in-context learning for code generation.
    \begin{figure*}[h]
        \centering
        \includegraphics[width=\linewidth]{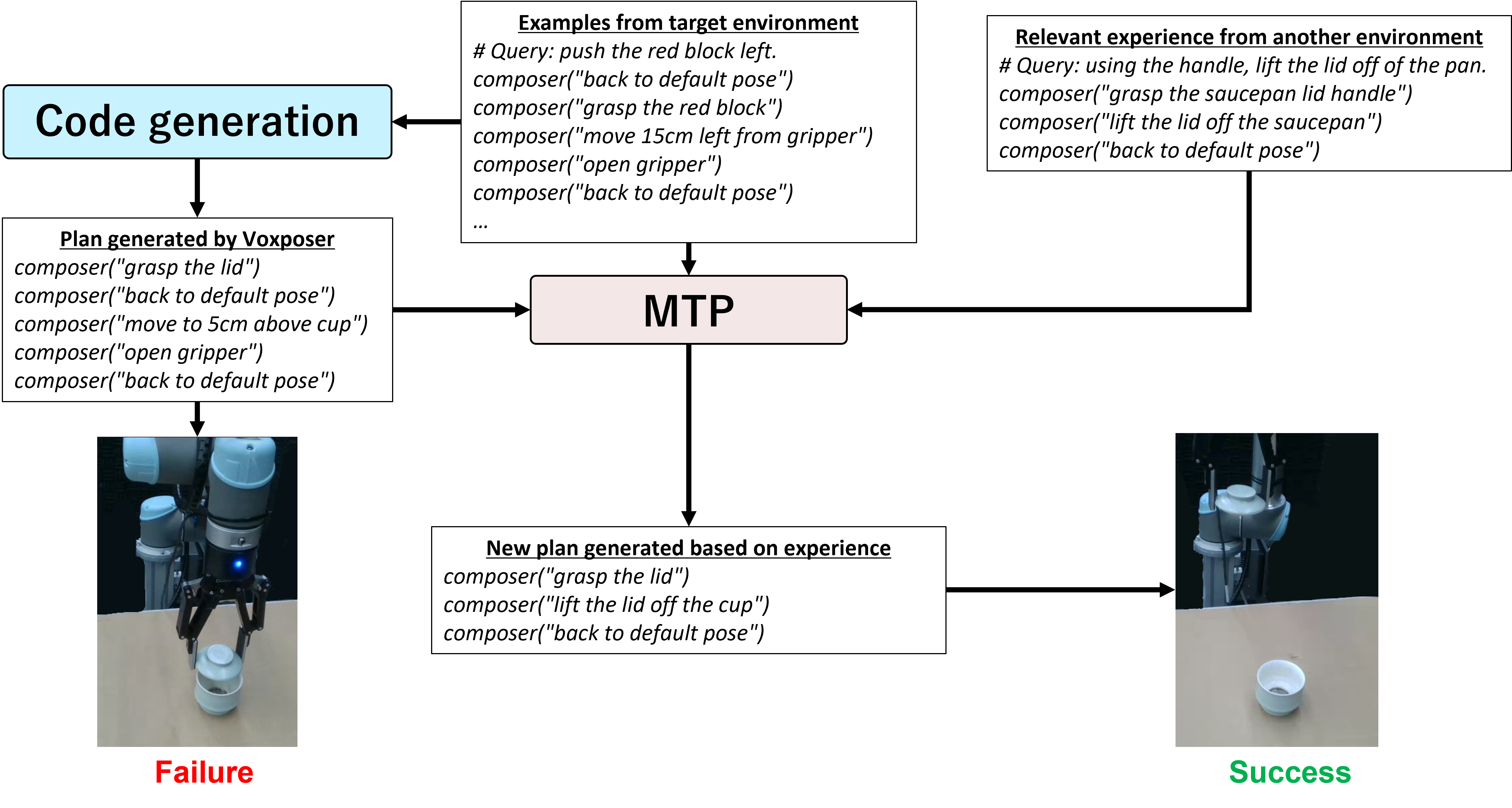}
        \vskip -0.1in
        \caption{Demonstration of memory-guided re-planning on UR5 for the ‘Take Lid Off Cup’ task — initial plan fails by resealing the cup, corrected plan succeeds by lifting the lid away.}
        \label{fig:ur5_lidtask}
    \end{figure*}

 \begin{table*}[h] 
        \caption{Success rate(\%) on RLBench.}
      \centering
      {\small
      \begin{minipage}{\textwidth}
        \centering
        \label{tab:rlbench-table}
        \begin{threeparttable}
          \begin{tabular}{c *{9}{c} c} 
            \toprule
             Method & BballHoop & Buzz & Drawer & LampOff & Bottle & Button & TrashBin & LidOff & Umbrella  &  mean ± std \\
            \midrule

            VoxPoser & 20.0 & 20.0 & 13.3 & 60.0 & 40.0 & 0.0 & 93.3 & 33.3 & 73.3 & 39.3 ± 3.4 \\

            Retry & \textbf{40.0} & 26.7 & 53.3 & 66.7 & 93.3 & 0.0 & 93.3 & \textbf{40.0} & 86.7 & 55.6 ± 6.7 \\

            Self-reflection & 26.7 & 20.0 & 80.0 & \textbf{73.3} & 86.7 & 33.3 & \textbf{100.0} & 33.3 & \textbf{93.3} & 60.7 ± 5.6 \\
            
            MTP(Ours) & 13.3 & \textbf{33.3} & \textbf{86.7} & 60.0 & \textbf{100.0} & \textbf{73.3} & \textbf{100.0} & 33.3 & 80.0 & \textbf{64.4} ± 2.2 \\

            \bottomrule
          \end{tabular}
        \end{threeparttable}
      \end{minipage}
      }
    \end{table*}

\subsubsection{Memory Adaptation}
\label{knowledge_transfer}

    Robot foundation models like RT-X \cite{embodimentcollaboration2024openxembodimentroboticlearning} demonstrate that shared information exists across diverse robotic tasks and environments. Building on this insight, we adapt retrieved code between environments, preserving essential planning elements while minimizing environment-specific differences.
    
    This process is illustrated in the left and center of Fig. \ref{fig:kt_figure}. Here, code examples from the target environment are provided as prompts, and the retrieved code is adapted to suit the target environment by LLMs.
    
    \begin{equation}
    \label{eqn:memory_adaptation}  
    \tilde c^{\text{tgt}} = \text{MemoryAdaptation}(c^{\text{src}}, P_{E^{\text{tgt}}} )
    \end{equation}
    
    As shown in (\ref{eqn:memory_adaptation}), the adapted code $\tilde{c}^{\text{tgt}}$ for the target environment is generated from the source code $c^{\text{src}}$ using a prompt $P_{E^{\text{tgt}}}$, where $P_{E}$ denotes a fixed prompt for the target environment $E^{\text{tgt}}$. This allows us to convert environment-specific plans (such as initializing a robot arm at the beginning) and environment-dependent information, like coordinates and scales, to match the target environment.

\subsubsection{Re-planning}
\label{re_planning}

    Using the code adapted by Knowledge Transfer, MTP proceeds with Re-planning. Fig. \ref{fig:replan_figure} shows this process using the examples, most relevant experience from memory and failed plan. If the initially generated code fails, the agent utilizes the most similar retrieved code in LLM inference context to generate new code. If this new attempt also fails, the agent uses the second most similar code to generate the code again. In this manner, MTP incorporates new ideas from similar knowledge to perform Re-planning.
    By combining initial prompt examples with those retrieved from memory, we leverage in-context learning. This approach allows us to maintain code quality while integrating new insights.

    \begin{equation}
    \label{eqn:replanning}
    c_{new}^{\text{tgt}} = \text{Re-planning}(c_{fail}, \tilde c^{\text{tgt}}, P_{E^{\text{tgt}}} )
    \end{equation}

    Here, $c_{\text{new}}^{\text{tgt}}$ is generated by LLMs using previous failed code $c_{\text{fail}}$ as a reference, and $\tilde{c}^{\text{tgt}}$ as a successful prior example in a similar context.
    This methodology is akin to the technique for generating new ideas discussed by \cite{lu2024aiscientistfullyautomated} and demonstrates that leveraging archived ideas or insights for the target task is a viable approach. Consequently, this method enhances the agents' capabilities, improving their flexibility and creativity in task resolution.


\section{Experiments}
\label{sec:experiments}

    To assess the effectiveness of MTP, we conducted experiments both in simulators (RLBench and CALVIN) and on a real robot.

    These benchmarks are chosen for their widespread use in embodied agent research and their near-real-world robotic settings.
    These benchmarks share common actions, such as pick and place, but they each include different variations in terms of objects and types of tasks.
    In the following section, we present the evaluation results, demonstrating how memory transfer—where knowledge gained in one environment is applied to an unfamiliar one—enhances the agent's efficiency and performance. This approach allows us to assess the agent's adaptability to new challenges while leveraging previously learned knowledge to solve novel tasks, closely mimicking real-world problem-solving scenarios for embodied AI.

    \begin{figure*}[h]
        \centering
        \includegraphics[width=\linewidth]{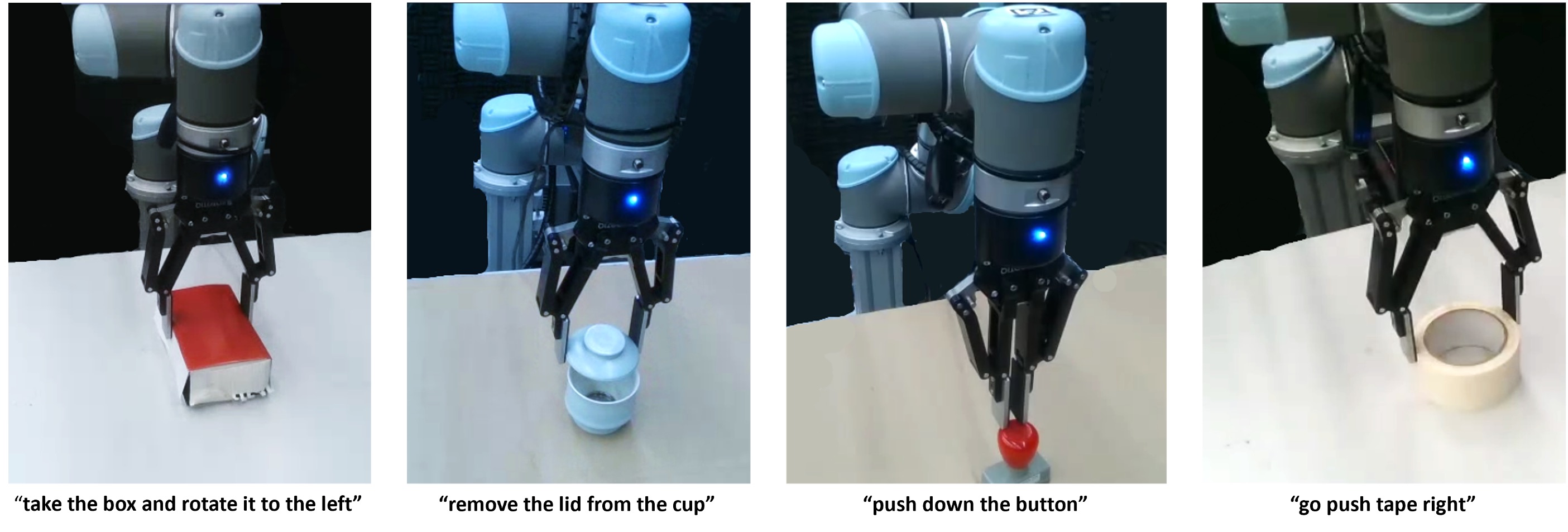}
        \vskip -0.1in
        \caption{UR5 executing various real-world tasks using MTP}
        \label{fig:ur5_tasks}
    \end{figure*}

\subsection{RLBench Environment}
\label{rlbench_experiments}

    RLBench \cite{james2019rlbenchrobotlearningbenchmark} is a robot learning benchmark for tabletop environments with various tasks. We sampled nine tasks from it and conducted evaluations covering various tasks, instructions, and objects. Each task was evaluated with 5 different initial states, using instructions randomly sampled from RLBench. The agent was allowed up to a maximum of 3 trials. GPT-4.1-mini was used for this evaluation.

    We compare MTP with VoxPoser (Code Generation baseline) and two Re-planning baselines: Retry and Self-reflection. 
    
    Retry indicates re-executing without making any changes if the previous trial fails. As the method for Self-reflection, we used a novel conventional approach \cite{shinn2023reflexionlanguageagentsverbal} to generate a plan for the next trial. Additionally, to extend Self-reflection to RLBench, we create the prompts ourselves and use the image obtained from RLBench as Observation to generate the next plan.
    
    In the proposed method, we conduct the evaluation using memory constructed from successful codes in the CALVIN benchmark \cite{mees2022calvinbenchmarklanguageconditionedpolicy}. While detailed settings for the CALVIN benchmark are described in Section \ref{calvin_result}, we used the successful codes from 50 tasks in CALVIN, executed with the Retry  and Self-reflection Re-planning methods, to build the memory.
    Moreover, for methods excluding the baseline, the success rate is calculated based on up to three trials.

    Table~\ref{tab:rlbench-table} shows that our method, MTP, outperforms both Retry and Self-reflection in overall success rate, achieving the highest total score of 64.4\%. Although individual task performance varies across methods, MTP consistently achieves high success rates across a broader range of tasks. Compared to other Re-planning methods, MTP demonstrates a clear advantage in tasks such as \textit{Button} and \textit{Drawer}, which we attribute to its ability to leverage prior experience from other environments, allowing for better generalization and adaptation. This transfer of memory enables MTP to succeed where Retry struggles, highlighting the strength of our approach in real-world generalization.

\subsection{CALVIN Environment}

\label{calvin_result}

    CALVIN \cite{mees2022calvinbenchmarklanguageconditionedpolicy} is a language-conditioned robotic manipulation benchmark. We selected 50 single tasks from its validation dataset, which includes various tasks that involve manipulating blocks, drawers, sliders, and lights by pushing, lifting, rotating, and toggling. The agent was allowed up to 3 trials per task to complete it. For evaluation, we compared MTP against CALVIN's multi-context imitation learning model (MCIL) \cite{lynch2021languageconditionedimitationlearning} and a Retry baseline. Using GPT-4.1-mini as our LLM, we conducted two evaluations: one using CALVIN's original instructions and another using five additional LLM-generated paraphrased instructions per task to test instruction robustness and repeated these experiments thrice.

    As shown in Table \ref{tab:calvin-table}, the results demonstrate that our method consistently outperforms both the Retry and Self-reflection approaches in the single instruction evaluation, showcasing its superior effectiveness in more straightforward scenarios. Importantly, when faced with the increased complexity of paraphrased instructions, all methods experience some degradation in performance. However, our method exhibits significantly greater robustness, maintaining a clear advantage over Retry and Self-reflection despite the linguistic variations. This suggests that our approach is better equipped to generalize across diverse and nuanced instruction forms, making it more reliable and adaptable in real-world environments where instructions may vary widely in phrasing and structure.

    \begin{table}[h]
      \centering
      {\small
    \begin{minipage}{0.5\textwidth}
        \centering
        \caption{Success rate(\%) on CALVIN.}
        \label{tab:calvin-table}
        \begin{threeparttable}
        \begin{tabular}{c c c}
        \toprule
        Method & \makecell{Single\\Instruction}  & \makecell{Paraphrased\\Instructions}  \\
        \midrule
        MCIL & 32.0 & - \\
        VoxPoser & 52.0 ± 2.0 & 47.3 ± 2.3 \\
        Retry & 59.3 ± 1.2 & 53.3 ± 2.3 \\
        Self-reflection & 62.7 ± 1.2 & 56.0 ± 5.3 \\
        MTP(Ours) & \textbf{67.3 ± 3.1} & \textbf{59.3 ± 3.1}\\
        \bottomrule
        \end{tabular}
        \end{threeparttable}
      \end{minipage}
      \hfill
      }
    \end{table}

    \begin{table*}[h]
    \centering
    \caption{Comparison of plans generated from VoxPoser and MTP across different tasks.}
    \label{tab:plan_comparison}
    \begin{tabular}{@{}p{0.46\textwidth}@{\hspace{0.06\textwidth}}p{0.46\textwidth}@{}}
    \toprule
    \textbf{VoxPoser} & \textbf{MTP (Ours)} \\
    \midrule
    
    \multicolumn{2}{@{}l}{\textbf{Task: Take the box and rotate it to the left}} \\
    \begin{minipage}[t]{\linewidth}
    \small
    \begin{flushleft}
    \texttt{composer("grasp the red box")} \\
    \texttt{composer("back to default pose")} \\
    \texttt{composer("rotate the gripper to the left")} \\
    \texttt{composer("back to default pose")}
    \end{flushleft}
    \end{minipage}
    &
    \begin{minipage}[t]{\linewidth}
    \small
    \begin{flushleft}
    \texttt{composer("back to default pose")} \\
    \texttt{composer("grasp the red box")} \\
    \texttt{composer("turn counterclockwise by 90 degrees")} \\
    \texttt{composer("open gripper")} \\
    \texttt{composer("back to default pose")}
    \end{flushleft}
    \end{minipage} \\
    \midrule
    
    \multicolumn{2}{@{}l}{\textbf{Task: Go push the tape left}} \\
    \begin{minipage}[t]{\linewidth}
    \small
    \begin{flushleft}
    \texttt{composer("push the circular tape to the left")} \\
    \texttt{composer("back to default pose")}
    \end{flushleft}
    \end{minipage}
    &
    \begin{minipage}[t]{\linewidth}
    \small
    \begin{flushleft}
    \texttt{composer("back to default pose")} \\
    \texttt{composer("grasp the circular tape")} \\
    \texttt{composer("move 5cm right from the gripper")} \\
    \texttt{composer("open gripper")} \\
    \texttt{composer("back to default pose")}
    \end{flushleft}
    \end{minipage} \\
    \midrule
    
    \multicolumn{2}{@{}l}{\textbf{Task: Remove the lid from the cup}} \\
    \begin{minipage}[t]{\linewidth}
    \small
    \begin{flushleft}
    \texttt{composer("grasp the lid")} \\
    \texttt{composer("back to default pose")} \\
    \texttt{composer("move to 5cm above the cup")} \\
    \texttt{composer("open gripper")} \\
    \texttt{composer("back to default pose")}
    \end{flushleft}
    \end{minipage}
    &
    \begin{minipage}[t]{\linewidth}
    \small
    \begin{flushleft}
    \texttt{composer("grasp the lid")} \\
    \texttt{composer("move gripper 10cm up")} \\
    \texttt{composer("back to default pose")}
    \end{flushleft}
    \end{minipage} \\
    \midrule
    
    \multicolumn{2}{@{}l}{\textbf{Task: Push down the button}} \\
    \begin{minipage}[t]{\linewidth}
    \small
    \begin{flushleft}
    \texttt{composer("move to the center of the red button")} \\
    \texttt{composer("close the gripper")} \\
    \texttt{composer("back to default pose")}
    \end{flushleft}
    \end{minipage}
    &
    \begin{minipage}[t]{\linewidth}
    \small
    \begin{flushleft}
    \texttt{composer("back to default pose")} \\
    \texttt{composer("close gripper")} \\
    \texttt{composer("move 5cm up from the red button")} \\
    \texttt{composer("move 5cm down from the red button")} \\
    \texttt{composer("back to default pose")} \\
    \texttt{composer("open gripper")}
    \end{flushleft}
    \end{minipage} \\
    \bottomrule
    \end{tabular}
    \end{table*}
    
\subsection{Physical Robot Evaluation}
\label{ur5_result}
To validate the effectiveness of our approach in real-world settings, we carried out experiments on an UR5 robot outfitted with a Robotiq 2F gripper. An Intel RealSense RGB-D camera \cite{intelRealsense2020} was used to provide visual input. Object segmentation was performed using LangSAM \cite{medeiros2025langsegmentanything} to identify target objects.

  \begin{table}[h]
        \caption{Task success rates on the UR5 robot comparing VoxPoser and our MTP method (successes out of 5 trials).}
        \label{tab:ur5_task_comparison}
        \centering
        {\small
        \begin{tabular}{lcc}
        \hline
        \textbf{Tasks} & \textbf{VoxPoser} & \textbf{MTP} \\
        \hline
        Rotate box     & 0/5 & 5/5 \\
        Move object        & 1/5 & 3/5 \\
        Remove lid    & 3/5 & 4/5 \\
        Push button        & 0/5 & 2/5 \\
        \hline
        \textbf{Overall Success Rate} & 30\% & 75\% \\
        \hline
        \end{tabular}
        }
    \end{table}
We evaluated a variety of manipulation tasks including pushing, rotating, and picking-and-place. Specifically, we tested four representative tasks: (i) \textit{rotating the box 90° to the left}, (ii) \textit{removing the lid of a cup} , (iii) \textit{pressing a button}, and (iv) \textit{pushing an object to the right} (as shown in Fig. \ref{fig:ur5_tasks}). During the initial runs with VoxPoser, failures often occurred due to incorrect or incomplete plans. However, our method was able to leverage memory from previous experiences in other environments to re-plan and successfully complete the tasks. For these experiments, we combined the memory of both RLBench and CALVIN. Each task was performed five times, and the results are presented in Table \ref{tab:ur5_task_comparison}. By retrieving the most relevant task and adapting the corresponding plan, the system was able to successfully complete real-world tasks without additional retraining. Table \ref{tab:plan_comparison} shows the plan generated by both methods. For instance, in the \textit{rotate the box} scenario, VoxPoser completes the rotation but leaves the object suspended rather than returning it to a grounded position. In contrast, our plan, grounded in experience, includes both the rotation and proper placement, ensuring task completeness. In pushing tasks, if a direct push proves ineffective due to physical constraints, our method intelligently adapts by switching to a grasp-and-place strategy—still achieving the intended outcome. Similarly, for tasks like \textit{remove the lid from the cup}, our system recognizes that only the removal is required and avoids unnecessary re-placement as shown in Fig. \ref{fig:ur5_lidtask}. In the \textit{push button} task, our model applies force from the correct angle—pushing from above—ensuring functional interaction based on prior knowledge. One of the key advantages of our approach is its ability to learn effectively in simulation environments and transfer that experience to real-world scenarios, making it highly suitable for applications where reliability and generalization are critical.


\section{Ablation study}
\label{sec:ablation}

\subsection{Memory Adaptation}

To further investigate the contribution of memory adaptation in our proposed method, we conducted an ablation study comparing MTP with and without memory adaptation. As shown in Table~\ref{tab:memtransfer-table}, the absence of memory transfer results in a substantial performance drop across both RLBench and CALVIN benchmarks.

    \begin{table}[h]
      \centering
      {\small
      \begin{minipage}{0.5\textwidth}
        \centering
        \caption{Comparison of memory adaptation.}
        \label{tab:memtransfer-table}
        \begin{threeparttable}
        \begin{tabular}{c c c}
        \toprule
        Method & RLBench  & CALVIN  \\
        \midrule
        Retry & 55.6 ± 6.7 & 59.3 ± 1.2 \\
        MTP w/o Memory Adaptation & 49.3 ± 3.1 & 60.0 ± 4.0 \\
        MTP & \textbf{64.4 ± 2.2} & \textbf{67.3 ± 3.1}\\
        \bottomrule
        \end{tabular}
        \end{threeparttable}
      \end{minipage}
      \hfill
      }
    \end{table}

Removing memory adaptation leads to a clear drop in success rate for both RLBench (64.4\% → 49.3\%) and CALVIN (67.3\% → 60.0\%). This confirms that our memory adaptation process—adapting previously successful code from other environments—is essential for generalization. Even when environments differ in tasks and objects, converting memory into context-aware plans enables more robust and effective planning.

These results collectively validate the core hypothesis of our work: the adaptation of memory from prior successful trials in related environments substantially improves both generalization and task completion efficiency, reinforcing the effectiveness of memory-based re-planning in embodied agent scenarios.

\subsection{Memory Variation}

\begin{table}[h]
  \centering
  {\small
  \begin{minipage}{0.5\textwidth}
    \centering
    \caption{Comparison by memory configurations.}
    \label{tab:memsrc-table}
    \begin{threeparttable}
    \begin{tabular}{lcc}
      \toprule
      & \multicolumn{2}{c}{Evaluation benchmark} \\
      \cmidrule(lr){2-3}
      Memory (built from) & RLBench & CALVIN \\
      \midrule
      No memory(Retry) & 55.6 ± 6 & 52.0 ± 2.0 \\
      RLBench & \textbf{68.9 ± 4.4} & \textbf{67.3 ± 3.1} \\
      CALVIN  & 64.4 ± 2.2          & 64.0 ± 2.0          \\
      \bottomrule
    \end{tabular}
    \end{threeparttable}
  \end{minipage}
  \hfill
  }
\end{table}
To examine the impact of different memory configurations, we compared MTP using memory constructed from either RLBench or CALVIN. As shown in Table~\ref{tab:memsrc-table}, memory built from RLBench consistently achieved higher performance—not only on RLBench tasks (68.9\%) but also on CALVIN tasks (67.3\%)—compared to using CALVIN memory.

This result suggests that memory quality and diversity significantly influence transfer effectiveness. RLBench, which contains a broader range of object manipulations and task types, may provide richer and more generalizable memory representations. In contrast, CALVIN memory, while still effective, appears less transferable across domains.

These findings highlight the importance of constructing high-coverage memory to support generalization in memory-based re-planning systems.


\section{Conclusion and Limitations}
\label{sec:conclusion}

We introduced Memory-Transfer Planning (MTP), a flexible framework for robotic planning that leverages prior successful trials by retrieving and adapting executable code. Through extensive experiments on RLBench, CALVIN, and real robot platforms, we demonstrated that MTP significantly improves task success rates and adaptability. Our results highlight the importance of both memory retrieval and contextual adaptation in achieving generalizable manipulation.

While MTP provides a practical and scalable module for bridging the gap between simulation and real-world scenarios, several challenges remain. Currently, the memory is statically constructed and lacks mechanisms for dynamic growth or pruning, which may limit scalability in long-term deployments. Moreover, MTP operates purely on textual representations (code), without integrating visual or multimodal inputs such as images or point clouds at various stages of execution.

Addressing these limitations—by incorporating dynamic memory management and multimodal grounding—presents promising directions for future research, enabling more adaptive and robust robotic agents in open-ended environments.





\bibliographystyle{ieeetr}  
\bibliography{example}       



\end{document}